\documentclass[sigconf]{acmart}
\usepackage{enumerate}

\AtBeginDocument{%
  \providecommand\BibTeX{{%
    \normalfont B\kern-0.5em{\scshape i\kern-0.25em b}\kern-0.8em\TeX}}}

\copyrightyear{2021} 
\acmYear{2021} 
\setcopyright{acmcopyright}\acmConference[CIKM '21]{Proceedings of the 30th ACM International Conference on Information and Knowledge Management}{November 1--5, 2021}{Virtual Event, QLD, Australia}
\acmBooktitle{Proceedings of the 30th ACM International Conference on Information and Knowledge Management (CIKM '21), November 1--5, 2021, Virtual Event, QLD, Australia}
\acmPrice{15.00}
\acmDOI{10.1145/3459637.3481948}
\acmISBN{978-1-4503-8446-9/21/11}

\begin{document}
\fancyhead{}

%% Use this when submitting an article to a sponsored event. You'll
%% receive a unique submission ID from the organizers
%% of the event, and this ID should be used as the parameter to this command.
%%\acmSubmissionID{123-A56-BU3}

%%
%% The majority of ACM publications use numbered citations and
%% references.  The command \citestyle{authoryear} switches to the
%% If you are preparing content for an event
%% sponsored by ACM SIGGRAPH, you must use the "author year" style of
%% the next command will enable that style.

%%
%% end of the preamble, start of the body of the document source.
% \begin{document}

%\title{SAR-Net: A Scenario-Aware Ranking Network for Personalized Recommendation in Hundreds of Travel Scenarios}
\title{SAR-Net: A Scenario-Aware Ranking Network for Personalized Fair Recommendation in Hundreds of Travel Scenarios}


%\author{Paper ID: 1567}
% \author{Qijie Shen}
% \affiliation{%

% \affiliation{%

\author{Qijie Shen}
\affiliation{%
  \institution{Alibaba Group}
  \city{Hangzhou}
  \country{China}
}
\email{qijie.sqj@alibaba-inc.com}

\author{Wanjie Tao}
\affiliation{%
  \institution{Alibaba Group}
  \city{Hangzhou}
  \country{China}
}
\email{wanjie.twj@alibaba-inc.com}

\author{Jing Zhang}
\affiliation{%
  \institution{The University of Sydney}
  \city{Darlington NSW 2008}
  \country{Australia}
}
\email{jing.zhang1@sydney.edu.au}

\author{Hong Wen}
\affiliation{%
  \institution{Alibaba Group}
  \city{Hangzhou}
  \country{China}
}
\email{qinggan.wh@alibaba-inc.com}

\author{Zulong Chen}
\affiliation{%
  \institution{Alibaba Group}
  \city{Hangzhou}
  \country{China}
}
\email{zulong.czl@alibaba-inc.com}

\author{Quan Lu}
\affiliation{%
  \institution{Alibaba Group}
  \city{Hangzhou}
  \country{China}
}
\email{luquan.lq@alibaba-inc.com}

% \affiliation{%

% \affiliation{%

% \affiliation{%

\renewcommand{\shortauthors}{Shen, et al.}

\begin{abstract}

The travel marketing platform of Alibaba serves an indispensable role for hundreds of different travel scenarios from Fliggy, Taobao, Alipay apps, etc. To provide personalized recommendation service for users visiting different scenarios, there are two critical issues to be carefully addressed. First, since the traffic characteristics of different scenarios, e.g., individual data scale or representative topic, are significantly different, it is very challenging to train a unified model to serve all. Second, during the promotion period, the exposure of some specific items will be re-weighted due to manual intervention, resulting in biased logs, which will degrade the ranking model trained using these biased data. In this paper, we propose a novel Scenario-Aware Ranking Network (SAR-Net) to address these issues. SAR-Net harvests the abundant data from different scenarios by learning users' cross-scenario interests via two specific attention modules, which leverage the scenario features and item features to modulate the user behavior features, respectively. Then, taking the encoded features of previous module as input, a scenario-specific linear transformation layer is adopted to further extract scenario-specific features, followed by two groups of debias expert networks, i.e., scenario-specific experts and scenario-shared experts. They output intermediate results independently, which are further fused into the final result by a multi-scenario gating module. In addition, to mitigate the data fairness issue caused by manual intervention, we propose the concept of Fairness Coefficient (FC) to measures the importance of individual sample and use it to reweigh the prediction in the debias expert networks. Experiments on an offline dataset covering over 80 million users and 1.55 million travel items and an online A/B test demonstrate the effectiveness of our SAR-Net and its superiority over state-of-the-art methods. SAR-Net has also been deployed in the online travel marketing platform of Alibaba and is serving hundreds of travel scenarios.

\end{abstract}





\maketitle

\section{Introduction}
In recent years, Recommender Systems (RS) have played an increasingly important role in e-commerce platforms \cite{wen2020entire,wen2021hierarchically,zhang2020empowering,sarwar2001item}. It not only boosts businesses via traffic utilization, but also greatly reduces the time cost for users to find items of interest. There are two main phases in a typical e-commerce RS, i.e., \emph{Matching} and \emph{Ranking}, where \emph{Matching} is able to retrieve several thousands of candidate items from hundreds of millions of items, while \emph{Ranking} is responsible for sorting these candidates according to specific metrics, such as click-through rate (CTR) \cite{xiao2017attentional,wang2017deep,2016Deep} or conversion rate prediction (CVR) \cite{ma2018entire,wen2020entire,2018Multi,wen2021hierarchically}. Sometimes, after \emph{Ranking}, there is an additional phase called \emph{Reranking}, which is sometimes utilized to adjust the ranking results based on some manually defined rules, especially during the promotion period, such as the Double-Eleven Shopping Festival in China. 

As a specific instance of RS, the Alibaba travel marketing platform aims to provide personalized recommendation service to users from hundreds of travel scenarios in Fliggy, Alipay, and Taobao apps. For providing preeminent service for users visiting different travel scenarios, there are two critical issues encountered in practice. On the one hand, due to the individual topic or data scale for each scenario, the data distribution among these travel scenarios is significantly different, resulting in the difficulty of training a unified model to serve all. We call it as the \emph{Multi-Scenario Modeling Issue}. On the other hand, to ensure the definiteness of traffic for certain important merchants and items during the promotion period, there is always manual intervention to adjust the ranking results in the \emph{Reranking} phase, which will make the real exposure traffic be biased towards those intervened merchants or items. Consequently, when training with the biased logs, the model in \emph{Ranking} phase will inevitably learn more information towards these overexposed intervened items, resulting in unexpected self-reinforcement in the ranked results, i.e., systematic discrimination of disadvantaged items. We call it as the \emph{Data Fairness Issue}.

% More than 90\% of users will behave in two or more scenarios while a item may reside in multiple scenarios. 

For the \emph{Multi-Scenario Modeling Issue}, there are typically three kinds of solutions: 1) training individual ranking model for each scenario \cite{zhou2018deep,zhou2019deep}; 2) training a unified ranking model with all scenario data; and 3) using a multi-task learning framework to output multiple prediction results simultaneously \cite{caruana1997multitask,misra2016cross,ma2018modeling}, and each result for each scenario. However, the first method has obvious shortcomings. First, as the number of scenarios increases, maintaining individual model for each scenario requires a huge amount of cost. Second, since the data scale from certain scenarios is small, it is difficult to train an excellent ranking model only leveraging data from their own data. Alternatively, the second type of methods try to utilize the data of all scenarios. However, since the traffic characteristics of different scenarios are significantly different, it is very challenging to train a unified model that performs well in all scenarios. The third type of methods focus on the multi-task learning approaches while trying to harvest all data by discovering the explicit relationships among different scenarios. And existing multi-task learning approaches adopt an early-sharing strategy by learning a shared feature embeddings among different tasks, followed by feeding them into individual task-specific sub-network, respectively. However, the traditional multi-task modeling methods ignore the modeling of user interest transfer across scenarios, which consequently are not able to predict user interest accurately. In addition, the importance of input information from different scenarios is inevitably different while previous methods do not capture it explicitly.

For the \emph{Data Fairness Issue}, recent related studies on the bias issue mainly focus on exposure bias \cite{wang2016learning,ovaisi2020correcting,2020SamWalker}, selection bias \cite{UVj2015Probabilistic,steck2013evaluation}, etc. However, to the best of our knowledge, there is no previous research on intervention bias in recommender systems. A straightforward way to handle this problem is to down-sampling the overexposed items caused by intervention. However, this method requires too much engineering tricks and manpower to manually adjust the data set, which are not applicable due to the frequently changing intervention rules, especially during promotion period. Additionally, over down-sampling the data set will inevitably lead to serious wastage of usable data.

%It can be seen from the above that the existing multi-scenario modeling method does not make full use of the data, nor does it model the relationship and differences between the scenarios, and at the same time does not consider the user's interest transfer between different scenarios. 
To address the above two issues, we propose a Scenario-Aware Ranking Network (SAR-Net) in this paper. The network structure is based on a multi-expert network. Two attention modules are employed to extract the user's cross-scenario interest considering the scenario features and item features, respectively. A scenario-wise linear transformation is devised to strengthen the important information for each individual scenario. The linear transformation uses an element-wise operation, which almost does not increase the overall parameters of the model, but can leverage the differences and commonness between scenarios. To address the data fairness issue caused by manual intervention, we propose a Fairness Coefficient to measure the importance of samples in the scenario. It acts as the weight of the sample in the loss function and a useful feature in the bias-expert net of the expert network, which will be removed when deployed online. In this way, the proposed SAR-Net can not only fully learn the differences and relationship between scenarios, but also reduce the impact of intervention bias.

%Our major contribution in this work is a model called SAR-Net. The model is based on the multi expert network structure, which strengthens the information importance of each scenario through network design, and uses the design of gate network and loss function to alleviate the exposure bias of marketing scenario. Experiments on Alibaba industrial data set show that SAR-Net model achieves the effect of state-of-the-art both online and offline.

The main contributions of this work are summarized as follows:

%\begin{itemize}
%\item We highlight the significance of homogeneous task multi scenario joint modeling, which is easier to maintain than single scenario modeling, and can make full use of the data between scenarios.
%\item We propose an A Scenario-Aware Ranking Network (SAR-Net) that serves all scenarios. The network structure is based on a multi-expert network. Two attention modules are employed to extract the user's cross-scenario interest transfer from the two dimensions of the scenario and the item, and then the use of scenario-wise linear transformation to strengthen the important information of different scenarios.
%\item We firstly proposed a fairness coefficient to measure the importance of samples in the scenario, as the weight of the sample in the loss function, and model it separately in the bias net of the expert network to reduce the impact of intervention bias on model learning.
%\item We evaluate SAR-Net on the industrial item dataset and deploy it in the travel marketing platform of Alibaba. SAR-Net has been serving all travel scenarios of Alibaba, and has brought more than 5\% click-through rate increase.
%\end{itemize}

\begin{itemize}
\item We propose a novel SAR-Net that can predict users' cross-scenario interest given the scenario features and item features and extract important scenario-specific information across different scenarios adaptively.
\item We investigate the data fairness problem caused by manual intervention in recommender systems and propose a simple yet effective solution through the design of network structure and loss function.
\item Evaluation on both the offline dataset and online A/B test demonstrate the superiority of the proposed SAR-Net over representative methods. SAR-Net has been serving all travel scenarios of Alibaba and brought more than 5\% CTR increase.
\end{itemize}

\section{Related Work}
Our proposed method specifically tackles the multi-scenario prediction problem and data fairness problem, so we briefly review the most related work from the following aspects: 1) Single-Scenario CTR Prediction, 2)Multi-Task Learning, 3)Bias and Unfairness in Recommender System.

% investigate two challenges in recommender system in the context of training a unified model to serve all sce

% tackles the multi-scenario 

% conversion rate prediction problem by employing the multi-task learning framework

% over the entire space. Therefore, we briefly review the most related

% work from the following two aspects: 1) conversion rate prediction

% and 2) multi-task learning.
\subsection{Single-Scenario CTR Prediction}
Existing CTR prediction works mainly focus on single scenario modeling from the following several aspects: 1) feature interaction (e.g., FM \cite{5694074}, deepFM \cite{guo2017deepfm}); 2) user historical behavior (e.g., DIN\cite{zhou2018deep}, DIEN \cite{zhou2019deep}); and 3) combining matching and ranking (e.g., DMR \cite{lyu2020deep}). 

Factorization Machine (FM) is proposed to model feature interactions explicitly, while previous generalized linear models such as Logistic Regression (LR) \cite{richardson2007predicting} and Follow-The-Regularized-Leader (FTRL) \cite{mcmahan2013ad} lack the ability to solve interaction issue. Wide\&Deep \cite{cheng2016wide} and DeepFM \cite{guo2017deepfm} combine wide part (low-order) and deep part (high-order) features to improve the performance. FmFM \cite{sun2021fm} makes each field feature have different embedding dimensions, so as to reduce the amount of model parameters and avoid over fitting problem. DIN \cite{zhou2018deep} utilizes the attention mechanism to capture relative interests from the user behavior sequence with regard to the candidate item. DIEN \cite{zhou2019deep} further uses a GRU structure to capture the evolution of user interest. Considering a single vector might be insufficient to capture complicated user patterns, DMIN \cite{xiao2020deep} models user's multiple interests by a special designed extractor layer. DSIN \cite{feng2019deep} introduces a hierarchical view of behavior sequence by dividing it into sessions. DMR \cite{lyu2020deep} considers the relevance between user and item to achieve better performance.

\subsection{Multi-Task Learning}

Multi-Task Learning (MTL) \cite{caruana1997multitask} aims to improve generalization by sharing knowledge across multiple related tasks. The shared knowledge and task-specific knowledge are explored to facilitate the learning of each task. There have been some studies applying the gate structure and attention network for information fusion. MOE \cite{jacobs1991adaptive} has a shared-bottom model structure, where the bottom hidden layers are shared across tasks. MMOE \cite{ma2018modeling} extends MOE to utilize different gates network for each task to obtain different fusion weights in MTL. Similarly, MRAN \cite{zhao2019multiple} applies multi-head self-attention to learn different
representation subspaces at different feature sets. Cross-Stitch \cite{misra2016cross} uses linear cross-stitch units to learn an optimal combination of task-specific representations.
% Specifically, each domain has a 7-layer fully-connected network and the cross-stitch units are added in each hidden layer to learn task-specific representations. 
To address the seesaw phenomenon, PLE \cite{tang2020progressive} separates shared components and task-specific components explicitly and adopts a progressive routing mechanism to extract and separate deeper semantic knowledge gradually. 

\subsection{Bias and Unfairness in RS}
User behavior data are observational rather than experimental. leading to various biases in the data, such as exposure bias, popularity bias, and unfairness bias \cite{chen2020bias}. Blindly fitting the data without considering the inherent biases will cause many serious issues, e.g., the discrepancy between offline and online performance, and reducing user's satisfaction and trust on the recommendation service. Exposure bias happens as users are only exposed to a part of specific items so that unobserved interactions do not always represent negative preference. Popularity bias can be explained as popular items are recommended even more frequently than their popularity would warrant \cite{abdollahpouri2020multi}. Unfairness bias can be explained as the system systematically and unfairly discriminates against certain individuals or groups of individuals in favor others \cite{Batya1996Bias,lambrecht2019algorithmic,2014Automated,stoica2018algorithmic,karimi2018homophily}. 

The intervention bias investigated in this paper shares similarity with unfairness bias, both of which are caused by the unbalanced data distribution. Consequently, a model trained on the data is biased. The difference is that unfairness bias refers to the imbalance of users, which misleads the model to lean towards the interests of specific users, which will affect the performance for long-tail users. The intervention bias causes the data fairness issue where external manual intervention adjusts the ranking results, biasing the model towards certain items or merchants. Trained on such unbalanced data, the model will overfit over-represented items and reinforce them in the ranked results, resulting in a systematic discrimination that reduces the visibility of disadvantaged items.

\section{Problem Formulation}
In a recommender system, the user-item interaction is typically formulated as a matrix $\textit{Y} = \{y_{ui}\}_{M\times N}$, where $M$ and $N$ denote the numbers of users and items, respectively. The interaction $y$ is either implicit feedback \cite{agarwal2009spatio}, e.g., click or explicit user rating \cite{koren2009collaborative}. In this work, we focus on the CTR prediction task, implying that the matrix $\textit{Y}$ consists of 0 and 1. Specifically, $\textit{y}_{uis} = 1$ means that user $u$ has clicked item $i$ in scenario $s$, otherwise $\textit{y}_{uis} = 0$. Moreover, each interaction is associated with a timestamp $t$ that records the time of interaction. Therefore, the data in the recommender system are denoted by a set of quintuplets $\Gamma = \{(u,i,s,t,y)\}$, each of which includes the user $u\in U$ interacts with an item $i\in I$ in scenario $s$ at a recommendation time $t$. For a target quadruple $(u,i,s,t)$, an interaction probability should be predicted. 

% The behavior history of users in the whole scenarios and intervention bias can be respectively defined as follows.

% Definition 1~(Cross-scenario User Behavior): Given a user $u$, the user behavior in all scenarios is a set of items that the user $u$ interacts in chronological order, i.e., 
% $\textit{I}_{u}^{t}=\{ i\mid (u,i,s,{t}')\in \Gamma ,{t}'<t\}=\{i_{1},i_{2}, \cdots \}$.

% Definition 2~(Intervention Bias): Due to the influence of intervention, the ranking results will be changed. Consequently, the data distribution for model training is also changed.

In this work, the CTR prediction model aims to estimate the probability of interaction $\hat{y}$ between a target user $u\in U$ and a candidate item $i\in I$ in scenario $s$, with the consideration of the scenario context and item context.

\section{Model Design}
\begin{figure*}
  \centering
  \includegraphics[width=\textwidth]{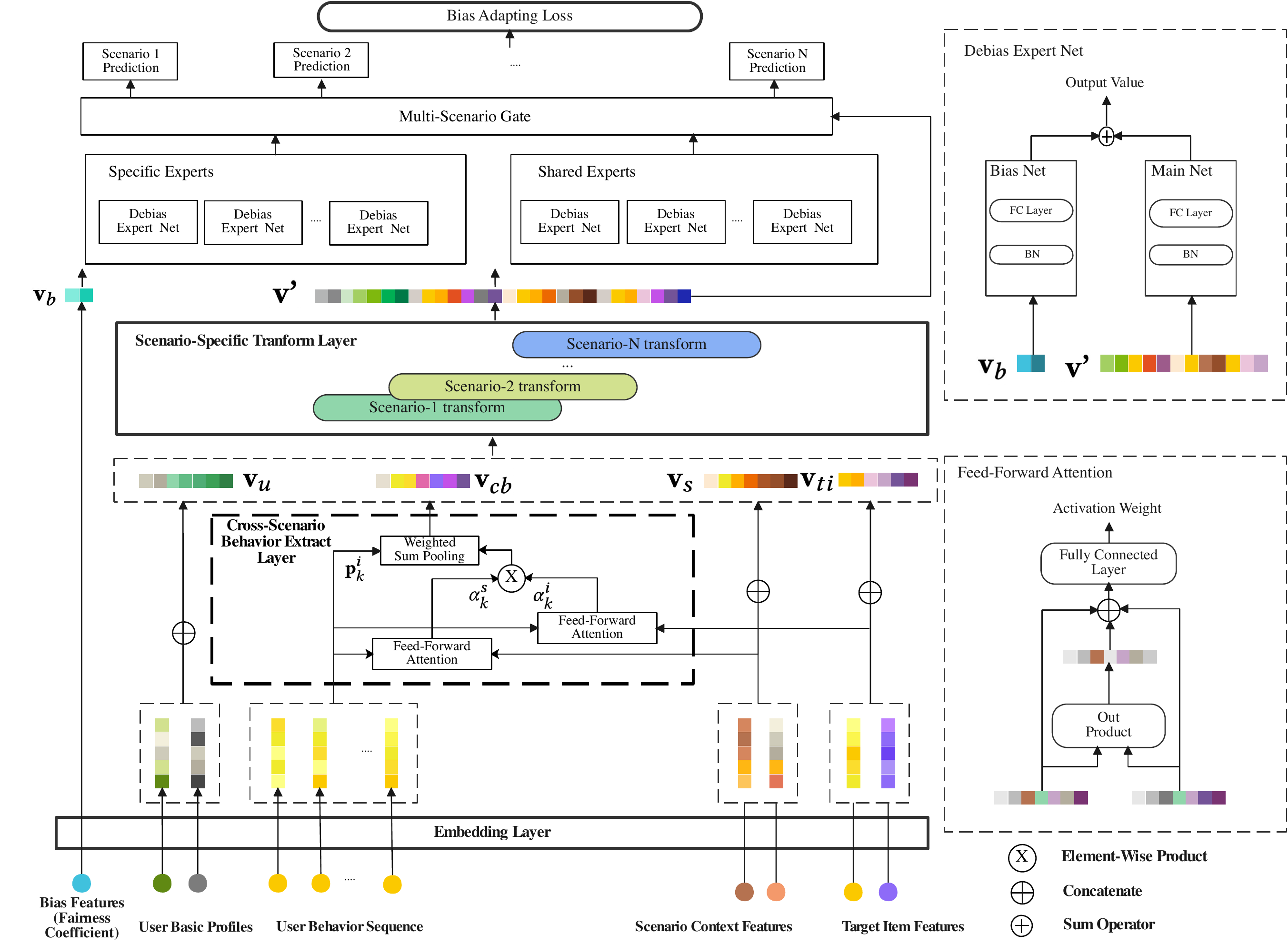}
  \caption{The overview architecture of our proposed model SAR-Net. Cross-Scenario Behavior Extract Layer harvests the abundant data from different scenarios by learning users' cross-scenario interest via two remarkable attention modules. Then, a Scenario-Specific Transform Layer is adopted to further extract scenario-specific information, followed by two groups of debias expert networks. We use a Fairness Coefficient to measure the importance of individual sample and use it to reweigh the prediction in the debias expert networks. A multi-scenario gating module is used to fuse these predictions into the final one.  }
  \label{fig:model_arc}
\end{figure*}

In this section we will detailedly present the SAR-Net, with its overall architecture illustrated in Figure \ref{fig:model_arc}. SAR-Net takes cross-scenario user behaviors, user basic profiles, contextual scenario features, and target item as input. It firstly embeds these input features as low-dimensional vectors by an embedding layer. Then, it extracts user cross-scenario interest transfer from users' historical behaviors by devising a Cross-Scenario Behavior Extract Layer after considering the scenario features and item features. Next, taking the encoded features of previous module as input, a scenario-wise linear transformation layer is adopted to strengthen the important information for each individual scenario. Finally, the Mixture of Debias Experts, i.e., scenario-specific experts and scenario-shared experts, output intermediate results independently,
which are further fused into the final result by a multi-scenario gating module. Now, we will introduce each module in detail.
  
\subsection{Fairness Coefficient}
Intervention bias makes the data distribution biased towards the weighted items. When training on such unbalanced data, the recommendation models will inevitably tend to learn more about these over-represented items, resulting in unexpected self-reinforcement in the ranked results, i.e., systematic discrimination of disadvantaged items. Therefore, to address this issue, we propose the concept of Fairness Coefficient (\emph{FC}), which measures the importance of individual sample and represents the degree of intervention of different items. 

Assuming $\mathbb{D}$, $d_{i,s}$, $N_{s}$ denotes the whole data of all scenarios in one day, the partial data of item $i$ exposing in scenario $s$, the number of items in scenario $s$, respectively, where $d_{i,s} \in \mathbb{D} $. Then, we define $PV(i,s)$ and $F(i,s)$ as the number of samples from $d_{i,s}$, the sum of the values predicted by the proposed SAR-Net for all samples from $d_{i,s}$ with an offline manner, respectively. Therefore, the Fairness Coefficient, denoted as  $w_{fairness}^{i,s}$, indicating the degree of intervention of different items in different scenarios, defined as follows:
\begin{equation}
\begin{split}
w_{fairness}^{i,s}=\frac{F(i,s)/\sum_{i=0}^{N_{s}}F(i, s)}{PV(i,s)/\sum_{i=0}^{N_{s}}PV(i, s)}. \\
\end{split}
\end{equation}

Where the numerator is a constant, which has no relationship with whether items intervened, while denominator affected by intervention. Therefore, as the increment of exposure volumes caused by intervention, the \emph{FC} will become smaller, vice versa. \emph{FC} will act as the weight of the sample in the loss function and a useful feature in the bias-expert net of the expert network.

\subsection{Embedding Layer}

There are five groups of features, i.e., user profiles, user cross-scenario behavior, scenario context feature, target item, and intervention bias. User profile contains features related to the user, e.g., user id, country, etc. Target item feature refers to the candidate item with corresponding features such as item id, category id, statistical offline scores, etc. Scenario context feature is a group of features including but not limited to time, current scenario id, current scenario type, etc. User cross-scenario behavior, with behavior type clicking, purchasing or add-to-cart, is a list of user interacted items in all scenarios, where each item in this list not only has same feature fields as the target item but also has the scenario context features at the moment that the behavior happened. Intervention bias feature is the Fairness Coefficient of each sample which is computed after one-day data is generated. Each feature can be encoded into an one-hot vector with high-dimension.

We first encode features into one-hot encodings. For the $i$th feature, its one-hot encoding is denoted as:
\begin{equation}
    \textrm{v}_{i} = onehot(i),
\label{eq:onehot}
\end{equation}
where $\textrm{v}_{i} \in \mathbb{R}^{N}$ is a vector with 1 at the $i$th entry and 0 elsewhere, and N is the number of unique features. We then map the sparse and high-dimensional one-hot encodings to dense and low-dimensional embedding vectors that are suitable for neural networks. In particular, we define a learnable embedding matrix $\textbf{E} \in \mathbb{R}^{D \times N}$, where $D\ll N$ is the dimension. The $i$th feature is then projected to its corresponding embedding vector $\textbf{e}_{i}\in\mathbb{R}^{D}$ as:
\begin{equation}
    \textbf{e}_{i}=\textbf{E}{v_{i}}.
\label{eq:embedding}
\end{equation}
%Equivalently, the embedding vector $\textbf{e}_{i}$ is the $i$th column of the embedding matrix $\textbf{E}$.

\subsection{Cross-Scenario Behavior Extract Layer}

Existing multi-task modeling methods do not consider the user's interest transfer in different scenarios. In fact, most users reside in more than one scenario, with different preference for respective scenario. For example, users prefer items such as tickets of scenic spots in the travel-surrounding theme scenario, while prefer items with higher living quality or with a place suitable for children to play in the parent-child theme scenario. And in the couple-travel theme scenario, romantic scenery and travel experience will become the primary consideration. Therefore, for reflecting user's significant travel intention for different scenarios and portraying user's interest transfer, the modeling for user's cross-scenario behaviors is very critical.

With this in mind, we can aggregate user's cross-scenario behaviors into a unified representation. In general, the aggregated strategy can be defined as:
\begin{equation}
\textit a_{s} = \sum_{i}^{}\alpha _{i}x_{i},
\end{equation}
where $\alpha_{i}$ is a weight assigned to $x_{i}$, indicating its importance during aggregation. The remaining issue is how to compute the weight. A naive way is $\alpha _{i} =1 / \left | x_{i}\right |$, i.e., each one of clicked items has equal importance. It is clearly not a wise choice because some items may not be indicative for the target item. Inspired by this insight, DIN \cite{zhou2018deep} applies the attention mechanism to extract relevant interests from user's behaviors, which considers the degree of correlation between historical behavioral items and target items. However, in multi-scenario modeling, the scenario context information at the moment that the behavior happened also carry a lot of information. For example, a user's historical behaviors in three themed scenarios, namely travel-surrounding, parent-child travel, and northwestern travel, provide informative context when users are looking for items they are interested in at the couple-travel theme scenario. The behavioral mentality of users in different scenarios in the past is different from that of the current scenario. Among them, the relevance of parent-child travel is weaker, therefore we need to consider the user's behavior in strong relevance scenarios to make recommendations. Specifically, we extract users' cross-scenario interests via two specific attention modules, which leverage the scenario features and item features to modulate the user behavior features, respectively.

Furthermore, a user's cross-scenario behavior can be split into two parts, i.e., item behavior sequence ${\textbf{p}}(B^{i})=\{\textbf{p}_{1}^{i},\textbf{p}_{2}^{i},\cdots ,\textbf{p}_{\left |{\textbf{p}}(B^{i}) \right |}^{i}\}$ and scenario context sequence ${\textbf{p}}(B^{s})=\{\textbf{p}_{1}^{s},\textbf{p}_{2}^{s},\cdots ,\textbf{p}_{\left |{\textbf{p}}(B^{s}) \right |}^{i}\}$, where $\textbf{p}_{k}^{i}$ is obtained by concatenating the \emph{k}th corresponding item feature embedding vectors, including item id, category, destination, etc, i.e., $\textbf{p}_{k}^{i} = [\textbf{e}_{itemId}||\textbf{e}_{destination}||\textbf{e}_{category}||\cdots  ]$. $\textbf{p}_{k}^{s}$ is obtained by concatenating the \emph{k}th corresponding scenario context feature embedding vectors, including scenario id, scenario type, behavior time, etc, i.e., $\textbf{p}_{k}^{s} = [\textbf{e}_{scenarioId}||\textbf{e}_{scenarioType}||\textbf{e}_{behaviorTime}||\cdots  ]$, where $||$ is the vector concatenation operator. 

Additionally, we define $\alpha_{k}^{i}$ and $\alpha_{k}^{s}$ , indicating the relevance between user's \emph{k}th behavior item and the target item or target scenario, respectively, shown as follows.

\begin{equation}
\begin{split}
\alpha_{k}^{i}=\frac{\exp(\psi(\textbf{p}_{k}^{i}, \textbf{p}_{t}^{i}))}{\sum_{l=1}^{\left | \textbf{p}(B^{i})\right |}\exp(\psi(\textbf{p}_{l}^{i}, \textbf{p}_{t}^{i}))}., \\
\end{split}
\end{equation}
\begin{equation}
\begin{split}
\alpha_{k}^{s}=\frac{\exp(\psi(\textbf{p}_{k}^{s}, \textbf{p}_{t}^{s}))}{\sum_{l=1}^{\left | \textbf{p}(B^{s})\right |}\exp(\psi(\textbf{p}_{l}^{s}, \textbf{p}_{t}^{s}))}, \\
\end{split}
\end{equation}

Where, $\textbf{p}_{t}^{i}$ and $\textbf{p}_{t}^{s}$ representing the embeddings of target item and target scenario, respectively. And $\psi(x,y)$,
taking two vectors $x$ and $y$ as input, output the weight value by employing the feed-forward attention operator, illustrated in Figure \ref{fig:model_arc}.

Finally, we aggregate $\textbf{p}_{k}^{i}$ with the consideration of both $\alpha_{k}^{i}$ and $\alpha_{k}^{s}$ as follows to get the user cross-scenario interest transfer $\textbf{v}_{cb}$.
\begin{equation}
\begin{split}
\textbf{v}_{cb} = \sum_{k=1}^{t}\alpha_{k}^{i}*\alpha_{k}^{s}* \textbf{p}_{k}^{i}.
\end{split}
\end{equation}

\subsection{Scenario-Specific Transform Layer}

After generating user interest transfer vector $\textbf{v}_{cb}$, user basic profiles vector $\textbf{v}_{u}$, target item feature vector $\textbf{v}_{ti}$ and the scenario context feature vector $\textbf{v}_{s}$, we can obtain the gathered representation \textbf{v} by concatenating the corresponding feature embedding vectors, i.e., $\textbf{v} = [\textbf{v}_{cb} || \textbf{v}_{u}] || \textbf{v}_{ti} || \textbf{v}_{s}]$. Next, for further extracting scenario-aware specific information, we apply a scenario-wise transform module to process the previous representation \textbf{v}. Specifically, for the $i$th scenario, we compute $\textbf{v'}$ as follows:
\begin{equation}
 \textbf{v'} = \textbf{v} \otimes \beta_{i}+\gamma _{i}.
\end{equation}
where, the vectors $\beta_{i}$ and $\gamma _{i}$ are scenario-aware parameters, which have the same dimension with $\textbf{v}$, and $\otimes$ is an element-wise operator.
%Scenario-specific transform layer adapts the shared embeding to the differences in different scenarios, and can extract important features in the scenario.

\subsection{Mixture of Debias Experts}
Because sharing the parameters among different tasks is difficult to describe the heterogeneity of different tasks and may potentially result in the negative transfer issue, we use a multi-expert network as the core structure of the feature extraction part. In order to further model the difference between homogeneous tasks, each scenario has some scenario-specific experts and all the scenarios share several common experts. Compared with the shared-only experts structure, our network alleviates the seesaw phenomenon, i.e., the model has a profit in one of the tasks but a negative profit in the other task.
%In addition, sharing and unique structure can make different expert networks more distinguishable.

In order to alleviate the influence of intervention bias on model prediction, we divide each expert network into two parts: Bias net and Main net. Both modules are composed of a fully connected network and a batch normalization layer.
Main net takes user interest transfer vector $\textbf{v}_{cb}'$, user basic profiles vector $\textbf{v}_{u}'$, target item feature vector $\textbf{v}_{ti}'$, and scenario context feature vector $\textbf{v}_{s}'$ as input, and aims to predict the click-through rate of users on the target item. Bias net receives the input of Fairness Coefficient $\textbf{v}_{b}$, taking one specific value from set $w$ according to scenario-aware and item-aware principle, and predicts the degree of intervention bias to reweigh the predicted score of main net. Bias net is used only during training and will be removed when deployed online.

\subsection{Multi-Gate Network \& Prediction}

After the Mixture of Debias Experts, we have the 
predicted scores from both scenario-specific experts and scenario-shared experts. Denoting $x$ as the input representation, $m_{k}$ as the quality of scenario $k$'s scenario-specific experts, and $m_{s}$ as the quality of scenario-shared experts, we have:
\begin{equation}
 S^{k}(x) = [o_{k,1},o_{k,2}, \cdots ,o_{k,m_{k}},o_{s,1},o_{s,2}, \cdots ,o_{s,m_{s}}]^{T}.
\end{equation}
The structure of the multi-gate network is based on a single-layer feed-forward network with a SoftMax activation function. It acts as a selector to calculate the weighted sum of the selected predicting scores. More precisely, the output of scenario $k$'s gating network is formulated as follows:
\begin{equation}
y^{k}(x) = w^{k}(x)S^{k}(x),
\end{equation}
which is indeed the final predicted score of scenario $k$.

% Each gating network can learn to “select” a subset of shared experts to use conditioned on the input example. This allows the modeling of complex interactions among heterogeneous variables.

\begin{table*}[]
\caption{The percentage of training dataset and average CTR of each scenario.}
\resizebox{\textwidth}{7mm}{
\begin{tabular}{lllllllllllllllllllll}
\toprule
   Scenario        & \#1     & \#2    & \#3    & \#4    & \#5    & \#6    & \#7    & \#8    & \#9    & \#10   & \#11   & \#12   & \#13   & \#14   & \#15   & \#16   & \#17   & \#18    & \#19   & \#20   \\
           \midrule
Percentage & 11.20\% & 2.01\% & 5.30\% & 3.61\% & 2.39\% & 7.52\% & 3.44\% & 12.17\% & 8.51\% & 1.39\% & 4.33\% & 7.21\% & 3.99\% & 1.49\% & 6.63\% & 8.52\% & 4.25\% & 0.15\% & 2.78\% & 3.11\% \\
CTR        & 5.22\%  & 2.14\% & 6.39\% & 3.52\% & 4.55\% & 6.83\% & 4.32\% & 5.01\% & 9.39\% & 1.23\% & 9.34\% & 3.61\% & 5.25\% & 2.21\% & 5.29\% & 6.52\% & 4.58\% & 3.66\%  & 4.79\% & 6.12\%\\
\bottomrule
\end{tabular}}
\label{pic:dataset_des}
\end{table*}

\begin{table}[]
\caption{The percentage of users that visited certain number of scenarios (NoS) in past 30 days.}
\begin{tabular}{lllllll}
\toprule
  NoS         & $<$2 & 2$\sim$4 & 4$\sim$6 & 6$\sim$8 & 8$\sim$10 & $>$10 \\
           \midrule
Percentage & 11.21\% & 13.40\%  & 25.30\%  & 32.10\%  & 11.52\%   & 6.47\%   \\
\bottomrule
\end{tabular}
\label{pic:user_act_scene}
\end{table}

\subsection{Bias Adapting Loss}

To mitigate the intervention bias issue, we propose the concept of Fairness Coefficient to measure the importance of each sample and hope that the proposed model can get more information from samples with high Fairness Coefficients, due to the fact that these type of samples are less intervened. Furthermore, we use the binary cross entropy with the consideration of Fairness Coefficient, defined as below:
\begin{equation}
Loss = \sum_{l=1}^{|L_{k}|}  w_{fairness}^{i,k} * loss_{k,l,i},
\end{equation}
where $loss_{k,l,i}$ is $l$th sample of scenario $k$ and item $i$. $w_{fairness}^{i,k}$ is the Fairness Coefficient of scenario $k$ and item $i$. $loss_{k,l,i}$ is computed as follows:
\begin{equation}
loss_{k,l,i} = -I_{k,l,i}\log p_{k,l,i} - (1-I_{k,l,i})log (1-p_{k,l,i}),
\end{equation}
where $p_{k,l,i}$ is the output of SAR-Net, $I_{k,l,i}$ is the label of $l$th sample of scenario $k$ and item $i$. $I_{k,l,i}=1$ indicates the current user will click the recommended item and 0 otherwise.

\section{Experiments}

To comprehensively evaluate the proposed SAR-Net, we conduct experiments to answer the following research questions: \\
\textbf{RQ1}: How does SAR-Net perform compared with state-of-the-art models for multi-task CTR prediction? \\
\textbf{RQ2}: How about training a single model for each scenario using its own data or training a unified model for all scenarios using all data compared with the proposed SAR-Net? \\
\textbf{RQ3}: How about the impact of each part on the overall model?

\subsection{Experimental Setups}
\subsubsection{Datasets}
Due to the lack of public datasets for the multi-scenario CTR prediction task, we use Alibaba production data containing user click behaviors on 20 scenarios to perform the offline evaluation. The dataset contains users’ logs from the travel platform of Alibaba in one month, which is collected from October 20th to November 20th, 2020, with the Double-Eleven Shopping Festival, one of the most important annual festivals of Alibaba Group, included during this period. The dataset is further organized into the training dataset and the testing dataset. The training dataset covers over 80 million users and 1.55 million travel items. Table \ref{pic:dataset_des} shows the percentage of training dataset and average CTR of each scenario. Table \ref{pic:user_act_scene} shows the percentage of users that visited certain number of scenarios in past 30 days. It is obvious that different scenario has different distribution, and nearly 90\% of users have visited multiple scenarios in past 30 days.

\subsubsection{Competitors}
We will compare our SAR-Net with multi-task models and single-scenario models. Since existing multi-task models are usually to model different tasks for the same scenario, we will adapt them to the multi-scenario CTR prediction task in this paper to model the same task (CTR) for different scenarios.

Specifically, \textbf{Multi-Task Models} include:
\begin{enumerate}[1)]
\item \textbf{Hard Parameter-Sharing}: HPS \cite{caruana1997multitask} is the most basic and commonly used MTL structure, where the parameters are straightforwardly shared between different tasks. 
\item \textbf{Cross-Stitch}: Cross-Stitch \cite{misra2016cross} proposes to learn static linear combinations to fuse representations of different tasks. 
\item \textbf{MMOE}: MMOE \cite{ma2018modeling} applies gating networks to combine bottom experts based on the input to handle the differences between tasks. 
\item \textbf{CGC}: Compared with MMOE, CGC  \cite{tang2020progressive} separates the expert layer into shared experts and unique experts, enabling different types of experts to concentrate on learning different knowledge efficiently without interference. 
\item \textbf{PLE}: Compared with CGC, PLE \cite{tang2020progressive} adopts a progressive routing mechanism to extract and separate deeper semantic knowledge gradually.
\end{enumerate}

and \textbf{Single-Scenario Models} include:
\begin{enumerate}[1)]
\item \textbf{Wide\&Deep}: Wide\&Deep \cite{cheng2016wide} combines LR (wide part) and DNN (deep part). 
\item \textbf{PNN}: PNN \cite{qu2016product} automatically learns feature representations and high-order feature interactions. 
\item \textbf{DIN}: Deep Interest Network \cite{zhou2018deep} models dynamic user interest based on historical behavior for CTR prediction.
\end{enumerate}

\subsubsection{Parameter Settings}
Adam \cite{kingma2014adam} is used as the optimizer with the learning rate of 0.001 for all methods and the batch size is 2048. For all methods, the truncation length of user behavior is 50. For SAR-Net, each scenario has 2 specific experts, and 8 experts are shared for each scenario. DIN, PNN and Wide\&Deep use single-scenario data and mix-scenario data for training respectively. We run each method 10 times and report the average results.

\subsubsection{Metrics}
(1) \textbf{AUC}: AUC denotes the Area Under the ROC Curve over the test set. It is a widely used metric for CTR prediction, which reflects the probability that a model ranks a randomly chosen positive instance higher than a randomly chosen negative instance. The larger AUC is, the better the CTR prediction model performs. It is noteworthy that a small improvement in AUC is likely to lead to a significant increase in online CTR \cite{zhou2018deep}. Concretely, we use the AUC of each scenario and overall AUC (mixing samples from all scenarios to calculate the overall AUC) as the metrics.
(2) \textbf{RelaImpr}: RelaImpr is introduced in \cite{yan2014coupled} to measure the relative improvement of a target model over a base model. Since the AUC of a random model is 0.5, RelaImpr is defined as: 
\begin{equation} 
RelaImpr= \left ( \frac{AUC(target\;model)-0.5}{AUC(base\;model)-0.5}-1  \right ) \times 100\% .
\label{eq:relaimpr}
\end{equation}

% \usepackage{multirow}
\begin{table*}[]
\caption{AUC of different models on the offline Alibaba production dataset. Single-Scenario denotes that models are trained using single scenario data independently. Mix-Scenario denotes that models are trained using all-scenario data. Multi-Scenario denotes that models are trained based on multi-task learning. ``Overall'' denotes that mixing samples from all scenarios to calculate AUC.}
\begin{tabular}{l|ccc|ccc|cccccc}
\toprule 
        & \multicolumn{3}{c|}{Single-Scenario}                                                & \multicolumn{3}{c|}{Mix-Scenario}                                                   & \multicolumn{6}{c}{Multi-Scenario Model}                                                                                                                                           \\
        \cline{2-13}
   Scenario     & \multicolumn{1}{c}{Wide\&Deep} & \multicolumn{1}{c}{PNN} & \multicolumn{1}{c|}{DIN} & \multicolumn{1}{c}{Wide\&Deep} & \multicolumn{1}{c}{PNN} & \multicolumn{1}{c|}{DIN} & \multicolumn{1}{c}{HPS} & \multicolumn{1}{c}{Cross-Stitch} & \multicolumn{1}{c}{MMOE} & \multicolumn{1}{c}{CGC} & \multicolumn{1}{c}{PLE} & \multicolumn{1}{c}{\textbf{SAR-Net}} \\
        \midrule 
\#1     & 0.6610                          & 0.6526                  & 0.6621                  & 0.6556                         & 0.6499                  & 0.6568                  & 0.6701                            & 0.6731                           & 0.6733                   & 0.6752                  & 0.6768                  & \textbf{0.6811}                      \\
\#2     & 0.6740                          & 0.6681                  & 0.6751                  & 0.6686                         & 0.6654                  & 0.6698                  & 0.6801                            & 0.6821                           & 0.6831                   & 0.6852                  & 0.6858                  & \textbf{0.6901}                      \\
\#3     & 0.6744                         & 0.6685                  & 0.6755                  & 0.6690                          & 0.6658                  & 0.6702                  & 0.6805                            & 0.6816                           & 0.6809                   & 0.6808                  & 0.6822                  & \textbf{0.6921}                      \\
\#4     & 0.6842                         & 0.6843                  & 0.6883                  & 0.6788                         & 0.6816                  & 0.683                   & 0.6903                            & 0.6952                           & 0.6958                   & 0.6961                  & 0.6968                  & \textbf{0.7053}                      \\
\#5     & 0.6660                          & 0.6661                  & 0.6701                  & 0.6606                         & 0.6634                  & 0.6648                  & 0.6721                            & 0.6742                           & 0.6745                   & 0.6755                  & 0.6769                  & \textbf{0.6829}                      \\
\#6     & 0.6748                         & 0.6749                  & 0.6789                  & 0.6694                         & 0.6722                  & 0.6736                  & 0.6809                            & 0.6828                           & 0.6838                   & 0.6855                  & 0.6861                  & \textbf{0.6902}                      \\
\#7     & 0.6745                         & 0.6746                  & 0.6786                  & 0.6691                         & 0.6719                  & 0.6733                  & 0.6806                            & 0.6816                           & 0.6822                   & 0.6818                  & 0.6827                  & \textbf{0.6936}                      \\
\#8     & 0.6876                         & 0.6901                  & 0.6977                  & 0.6796                         & 0.6824                  & 0.6838                  & 0.6911                            & 0.6915                           & 0.6959                   & 0.6962                  & 0.6975                  & \textbf{0.7085}                      \\
\#9     & 0.6440                          & 0.6441                  & 0.6481                  & 0.6386                         & 0.6414                  & 0.6428                  & 0.6501                            & 0.6551                           & 0.6593                   & 0.6552                  & 0.6568                  & \textbf{0.6681}                      \\
\#10    & 0.6840                          & 0.6841                  & 0.6881                  & 0.6786                         & 0.6814                  & 0.6828                  & 0.6901                            & 0.6921                           & 0.6935                   & 0.6962                  & 0.6958                  & \textbf{0.7001}                      \\
\#11    & 0.6745                         & 0.6746                  & 0.6786                  & 0.6691                         & 0.6719                  & 0.6733                  & 0.6806                            & 0.6815                           & 0.6811                   & 0.6809                  & 0.6829                  & \textbf{0.6929}                      \\
\#12    & 0.6942                         & 0.6943                  & 0.6983                  & 0.6888                         & 0.6916                  & 0.6930                   & 0.7003                            & 0.7052                           & 0.7058                   & 0.7061                  & 0.7068                  & \textbf{0.7154}                      \\
\#13    & 0.6460                          & 0.6461                  & 0.6501                  & 0.6406                         & 0.6434                  & 0.6448                  & 0.6521                            & 0.6542                           & 0.6545                   & 0.6555                  & 0.6569                  & \textbf{0.6729}                      \\
\#14    & 0.6742                         & 0.6743                  & 0.6783                  & 0.6688                         & 0.6716                  & 0.6730                   & 0.6803                            & 0.6822                           & 0.6831                   & 0.6857                  & 0.6862                  & \textbf{0.6905}                      \\
\#15    & 0.6748                         & 0.6749                  & 0.6789                  & 0.6694                         & 0.6722                  & 0.6736                  & 0.6809                            & 0.6821                           & 0.6831                   & 0.6819                  & 0.6825                  & \textbf{0.6934}                      \\
\#16    & 0.6950                          & 0.6951                  & 0.6991                  & 0.6896                         & 0.6924                  & 0.6938                  & 0.7011                            & 0.7015                           & 0.7059                   & 0.7062                  & 0.7075                  & \textbf{0.7185}                      \\
\#17    & 0.6860                          & 0.6861                  & 0.6901                  & 0.6806                         & 0.6834                  & 0.6848                  & 0.6921                            & 0.6924                           & 0.6979                   & 0.6972                  & 0.6985                  & \textbf{0.7095}                      \\
\#18    & 0.6350                          & 0.6351                  & 0.6391                  & 0.6357                         & 0.6371                  & 0.6411                  & 0.6411                            & 0.6551                           & 0.6583                   & 0.6592                  & 0.6595                  & \textbf{0.6691}                      \\
\#19    & 0.6840                          & 0.6841                  & 0.6881                  & 0.6786                         & 0.6814                  & 0.6828                  & 0.6901                            & 0.6921                           & 0.6935                   & 0.6971                  & 0.6988                  & \textbf{0.7021}                      \\
\#20    & 0.6345                         & 0.6346                  & 0.6386                  & 0.6291                         & 0.6319                  & 0.6333                  & 0.6406                            & 0.6465                           & 0.6511                   & 0.6509                  & 0.6539                  & \textbf{0.6729}                      \\
Overall & 0.6640                          & 0.6642                  & 0.6682                  & 0.6587                         & 0.6615                  & 0.6629                  & 0.6702                            & 0.6759                           & 0.6762                   & 0.6791                  & 0.6801                  & \textbf{0.6997}       \\       \bottomrule        
\end{tabular}
\label{pic:exp_result}
\end{table*}

\subsection{Experimental Results: RQ1 and RQ2}
As illustrated in Table \ref{pic:exp_result}, the consistent improvement of our SAR-Net over different contenders validates its efficacy. Note that the overall performance of Mix-Scenario models are worse than the single-scenario model and the multi-scenario models, which proves obscuring scenario difference hurts the modeling of multi-scenario CTR prediction. Besides, HPS, Cross-Stitch, MMOE, CGC, and PLE all achieve better overall performance
than single-scenario methods and mix-scenario methods, demonstrating the importance of exploiting the distinctions and relationship between scenarios.

Although single-scenario methods all achieve better performance than mix-scenario methods, it is notable that in scenario \#18, the AUCs of single-scenario methods are worse than mix-scenario methods. We suspect that it is because the data of scenario \#18 which is 0.15 percent of training dataset as illustrated in Figure \ref{pic:dataset_des}, is not sufficient to train a reasonable model, while mix-scenario method has enough data to make it. On the other hand, in scenario \#8, the DIN model trained in the single-scenario manner achieves better performance than HPS, Cross-Stitch, MMOE, CGC, and PLE, which is opposite to the overall performance results, it shows that in some scenarios, existing multi-task models fail to extract deeper information about the scenarios and users. In contrast, the proposed SAR-Net exhibits superior performance across all scenarios compared with single-scenario methods and mix-scenario methods. Besides, SAR-Net also achieves consistent improvement than HPS, Cross-Stitch, CGC, and PLE, which shows the superiority of explicitly modeling the user's interest transfer and extracting scenario-specific information.

\begin{table}[]
\caption{Ablation study of the Cross-Scenario Behavior Extract Layer. SAR-Net$^{*}$ is the base model by replacing the Cross-Scenario Behavior Extract Layer with mean pooling.}
\begin{tabular}{lcc}
\toprule
                        & Overall AUC & RelaImp \\
                        \midrule
SAR-Net$^{*}$                 & 0.6925      & 0       \\
SAR-Net$^{*}$+Target Attention             & 0.6956      & 1.610    \\
SAR-Net$^{*}$+Scenario Attention       & 0.6934      & 1.004   \\
SAR-Net$^{*}$+Concatenate Attention   & 0.6944      & 0.987   \\
SAR-Net$^{*}$+Hierarchical Attention & 0.6955      & 1.558   \\
\textbf{SAR-Net}                 & \textbf{0.6997}      & \textbf{3.741}  \\
\bottomrule
\end{tabular}
\label{table:ablation_attention}
\end{table}

\begin{table}[]
\caption{Ablation study of the Bias Net and Bias Adapting Loss. SAR-Net$^{\dagger}$ denotes the base model that removes the bias net and uses the naive binary cross-entropy loss.}
\begin{tabular}{lcc}
\toprule
                          & Overall AUC & RelaImp \\
\midrule
SAR-Net$^{\dagger}$               & 0.6911      & 0       \\
SAR-Net$^{\dagger}$+sub-sampling(0.9)        & 0.6915      & 0.209   \\
SAR-Net$^{\dagger}$+sub-sampling(0.8)        & 0.6921      & 0.523     \\
SAR-Net$^{\dagger}$+sub-sampling(0.7)        & 0.6909      & -0.104   \\
SAR-Net$^{\dagger}$+sub-sampling(0.6)        & 0.6899      & -0.628  \\
SAR-Net$^{\dagger}$+Bias Net           & 0.6955      & 2.302   \\
SAR-Net$^{\dagger}$+Bias Adapting Loss & 0.6964      & 2.773   \\
\textbf{SAR-Net}                   & \textbf{0.6997}      & \textbf{4.500}  \\
\bottomrule
\end{tabular}
\label{table:ablation_bias}
\end{table}

\begin{table}[]
\caption{Ablation study of the Scenario-Specific Transform Layer. SAR-Net$^{\ddagger}$ denotes the base model without the Scenario-Specific Transform Layer.}
\begin{tabular}{lcc}
\toprule
                          & Overall AUC & RelaImp \\
\midrule
SAR-Net$^{\ddagger}$                  & 0.6948      & 0       \\
SAR-Net$^{\ddagger}$+multi-layer experts(2)        & 0.6951      & 0.256   \\
SAR-Net$^{\ddagger}$+multi-layer experts(3)        & 0.6956      & 0.523     \\
SAR-Net$^{\ddagger}$+multi-layer experts(4)        & 0.6958      & 0.359   \\
SAR-Net$^{\ddagger}$+multi-layer experts(5)        & 0.6957      & 0.301  \\
\textbf{SAR-Net}                   & \textbf{0.6997}      & \textbf{2.515}  \\
\bottomrule
\end{tabular}
\label{table:ablation_multilayer}
\end{table}

\subsection{Ablation Study: RQ3}
\subsubsection{Cross-Scenario Behavior Extract Layer}
In this section, we investigate the impact of the attention mechanism in the Cross-Scenario Behavior Extract Layer. The base model is SAR-Net$^{*}$, which extracts user behavior using a mean pooling operator (No Attention). In particular, we consider the following settings: 1) Target Attention: taking the target item as the query and the behavior item as the key in the attention mechanism; 2) Scenario Attention: taking scenario context feature as the query and the scenario context information at the moment that the behavior happened as key; 3) Concatenate Attention: concatenating target item feature and scenario context feature as query; 4) Hierarchical Attention: using a two-layer attention. In the first layer, scenario context feature concatenated with the target item is used as query. In the second layer, scenario context feature is used as the query. Multiplication of the weights of the two attention layers is used to generate the final pooling weights; and 5) Attention mechanism used in SAR-Net: target item and scenario context feature are respectively used as queries to learn the two-layer weights and element-wise item is used to generate the final pooling weights. 

As illustrated in Table \ref{table:ablation_attention}, ``No attention'' perform the worst, showing that useful signals could be easily buried in noise without distilling. In addition, either target item attention or scenario attention can improve the AUC compared with the base model, demonstrating that considering the relevance of target item or  scenario context information both can bring gains. The third and fourth attention mechanisms that consider both target item and scenario context features perform better than the base model but worse than target attention, implying that directly concatenating target item and scenario context feature as the query could not fully attend and exploit the useful information. In contrast, SAR-Net learns attention weights from the perspective of target item and  scenario context respectively and achieves the best performance. It shows that learning weights separately will avoid mutual interference, and can extract users' interest transfer across different scenarios.

\subsubsection{Bias Net and Bias Adapting Loss}
In this section, we investigate the impact of the Bias Net and Bias Adapting Loss in SAR-Net, which are used to mitigate the intervention bias issue. In particular, we try the loss without the weight of fair coefficients and the model without the bias net structure in the Expert Net as the base model, called SAR-Net$^{\dagger}$. We consider the following Settings: 1) sub-sampling items that are overexposed due to manual intervention. Sampling ratios were 0.9,0.8,0.7,0.6; 2) introducing the bias net to the expert network; 3) using the bias adapting loss; and 4) using the bias net and bias adapting loss. 

As can be seen from Table \ref{table:ablation_bias}, when the biased items being manual intervened are sub-sampled, the performance was improved slightly at first and then decreased. When the sub-sampling ratio is 0.7, the performance is the best. This is because sub-sampling will reduce the proportion of intervened items in the dataset, so that the model can learn the information of each type of items in a more fair way. However, when the proportion gradually decreases, the performance of the model will decrease, showing that excessive sub-sampling leads to the reduction of samples and the interaction information between users and items is not fully utilized. Bias net that is introduced to each expert net to reweigh the prediction of expert during training achieves better performance compared with base model and the sub-sampling method. Besides, bias adapting loss can make the model adaptive to the biased data and learn information according to the importance of samples. Our SAR-Net with both bias net and bias adapting loss can further boost the performance, demonstrating the effectiveness of the Bias Adapting Loss and Bias Net and their complementarity.

\subsubsection{Scenario-Specific Transform Layer}
Scenario-specific transform layer further models the differences and relations between scenarios by strengthening the key information of different scenarios. In this section, we take SAR-Net without the scenario-wise transform layer as the base model SAR-Net$^{\ddagger}$ and consider the following settings: 1) referring to PLE \cite{tang2020progressive}, a multi-layer expert network is used to extract features. And the number of layers is set to 2, 3, 4, 5, respectively; and 2) a single layer expert network is used and the scenario-wise transform layer is added before the expert network (SAR-Net). It can be seen from Table \ref{table:ablation_multilayer} that the benefit of the multi-layer extraction network structure converges gradually with the increase of the number of layers. Compared with the multi-layer expert extraction structure, scenario-specific transform layer achieves better results with less parameters.

\subsection{Online Deployment Test}

\begin{figure}
  \centering
  \includegraphics[width=\linewidth]{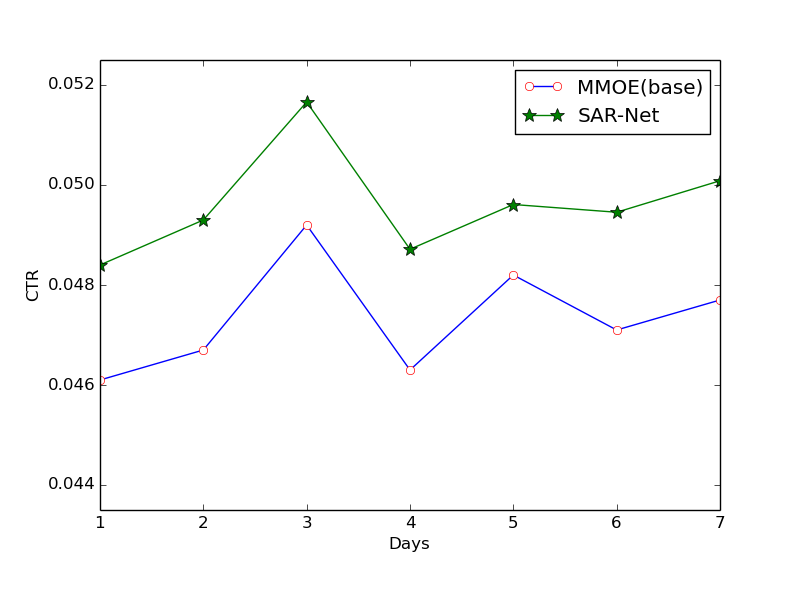}
  \caption{Online CTRs of SAR-Net and the base model in seven days in November 2020.}
  \label{pic:online_ctr}
\end{figure}

\begin{figure}
  \centering
  \includegraphics[width=\linewidth]{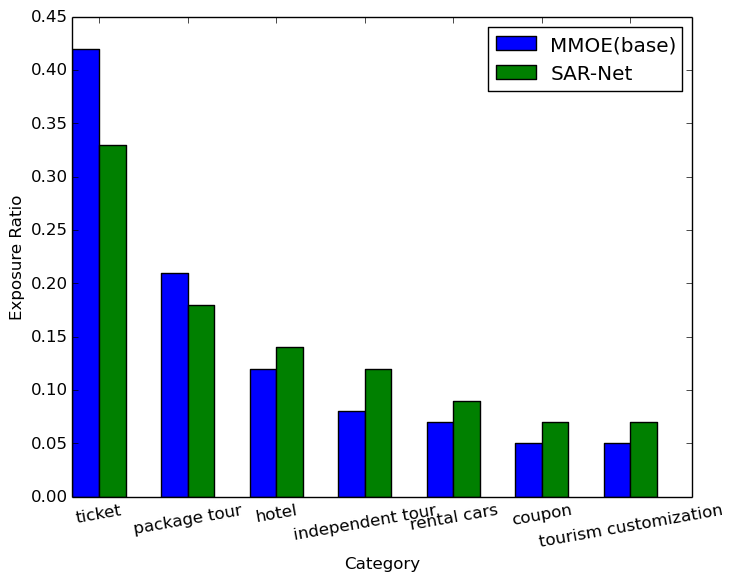}
  \caption{Online exposure ratios of SAR-Net and the base model as per category in seven days in November 2020. The intervened samples are removed from the statistics. Exposure ratio denotes the number of category's exposure over the number of all test data.}
  \label{pic:cate_pvratio}
\end{figure}

\begin{figure}
  \centering
  \includegraphics[width=\linewidth]{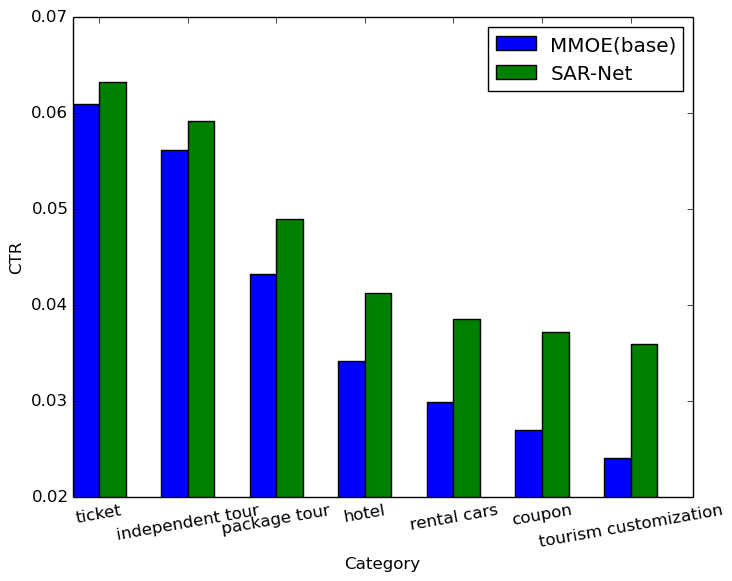}
  \caption{Online CTRs of SAR-Net and the base model as per category in seven days in November 2020. The intervened samples are removed from the statistics.}
  \label{pic:cate_ctr}
\end{figure}

We conduct the online A/B test by deploying our SAR-Net to handle real traffic in the personalized scenarios of Fliggy, Taobao and Alipay for seven days in November 2020, where the base model is MMOE \cite{ma2018modeling}. The online evaluation metric is real CTR, which defined as the number of clicks over the number of item impressions. The experimental results are shown in Figure \ref{pic:online_ctr}. It is clear that SAR-Net outperforms the base model MMOE \cite{ma2018modeling} consistently, demonstrating the effectiveness of SAR-Net in practical multi-scenario CTR tasks. SAR-Net has been deployed in the online travel marketing platform in Alibaba and is now serving hundreds of travel scenarios.

Moreover, we analyzed the performance of SAR-Net as per different categories compared with the base model. As illustrated in Figure \ref{pic:cate_pvratio} and Figure \ref{pic:cate_ctr}. We found that: 1) SAR-Net makes the exposure ratio of each category more even; 2) SAR-Net achieves consistent improvement in all categories; and 3) the improvement is more obvious on the categories with smaller traffic. These results demonstrate that SAR-Net effectively mitigates the intervention bias issue and achieves fair recommendation for each item.

\section{Conclusion}
In this paper, we propose a novel Scenario-Aware Ranking Network (SAR-Net) to address two issues encountered in the context of Alibaba travel marketing platform, i.e., multi-scenario modeling issue and data fairness issue. SAR-Net harvests the abundant data from different scenarios by learning users' cross-scenario interests via two specific attention modules. Then, a scenario-specific transformation layer is adopted to further extract scenario-specific features, followed by two groups of debias expert networks. Furthermore, above intermediate results are fused into the final result by a multi-scenario gating module. In addition, we propose the concept of Fairness Coefficient to measure the importance of individual sample and use it to reweigh the prediction in the debias expert networks. In this way, SAR-Net can address above two issues efficiently. The experimental results on both offline dataset and from online A/B test demonstrates the superiority of SAR-Net over representative methods for multi-scenario prediction. SAR-Net has been deployed in the online travel marketing platform of Alibaba and is serving hundreds of travel scenarios, bringing a 5\% improvement on CTR. In the future, we intend to investigate the impact of introducing more user fine-grained behaviors.

% SAR-Net harvests the abundant data from different scenarios by learning users' cross-scenario interests via two remarkable attention modules, which leverage the scenario features and item features to modulate the user behavior features, respectively. Then, taking the representation features of previous module as input, a scenario-specific linear transformation layer is adopted to further extract scenario-specific features, followed by two groups of debias expert networks, i.e., scenario-specific experts and scenario-shared experts. They output intermediate results independently, further be fused into the final result by a multi-scenario gating module. In addition, to mitigate the data fairness issue caused by manual intervention, we propose the concept of Fairness Coefficient (FC) to measures the importance of individual sample and use it to reweigh the prediction in the debias expert networks. The experiments demonstrates the superiority of SAR-Net on multi-scenario prediction and can effectively alleviate the intervention bias issue. 

% \begin{acks}
% \end{acks}

%\newpage
\bibliographystyle{ACM-Reference-Format}
\bibliography{CIKM2021_SARNet}

\appendix

\end{document}